\documentclass{article}
\usepackage[a4paper, margin=1in]{geometry}

\usepackage{booktabs} 
\usepackage{amsmath}
\usepackage{amssymb}
\usepackage{txfonts}
\usepackage{graphicx}
\usepackage{fancyhdr}
\usepackage[ruled]{algorithm2e} 

\begin{document}

\title{2-D Embedding of Large and High-dimensional Data with Minimal
Memory and Computational Time Requirements}

\date{}

\author{\textsc{Witold Dzwinel}\\
University of Science \& Technology, Department of Computer Science, Poland\\
dzwinel@agh.edu.pl\\
\\ 
\textsc{Rafa\l{} Wcis\l{}o}\\
University of Science \& Technology, Department of Computer Science, Poland\\
wcislo@agh.edu.pl\\
\\
\textsc{Stan Matwin}\\
Dalhousie University, Faculty of Computer Science, Canada\\
stan@cs.dal.ca
}




\newcommand{\mb}{\mathbf}

\maketitle

\begin{abstract}

In the advent of big data era, interactive visualization of large
	data sets consisting of $M \sim 10^{5+}$ high-dimensional feature
	vectors of length $N$ ($N\sim 10^{3+}$), is an indispensable tool for data exploratory
	analysis. 
The state-of-the-art data embedding (DE) methods of
	$N$-D data into 2-D (3-D) visually perceptible space (e.g., based on
	t-SNE concept) are too demanding computationally to be
	efficiently employed for interactive data analytics of large
	and high-dimensional datasets. 
Herein we present a simple method, {\bf ivhd} ({\bf i}nteractive {\bf v}isualization of
	{\bf h}igh-dimensional {\bf d}ata tool), which radically outperforms the modern 
	data-embedding algorithms in both computational and memory loads,
while retaining high quality of $N$-D data
	embedding in 2-D (3-D). 
We show that DE problem is equivalent to the nearest neighbor $nn$-graph 
	visualization, where only indices of a few nearest neighbors of each 
	data sample has to be known, and binary distance between data samples -- 0 to the
	nearest and 1 to the other samples -- is defined. 
These improvements reduce the time-complexity and memory load
	from $O(M\log M)$ to $O(M)$, and ensure minimal $O(M)$
	proportionality coefficient as well. 
We demonstrate high efficiency, quality and robustness of {\bf ivhd} on popular benchmark
	datasets such as MNIST, 20NG, NORB and RCV1.
\end{abstract}

%
%


\subsubsection*{Keywords}
Large high dimensional data, data embedding, $k$-$nn$ graph visualization, interactive data visualization


\section{Introduction}

In the age of data science interactive visualization of large high-dimensional
	datasets is an essential tool in exploratory data analysis and
	knowledge extraction. 
It allows for a better insight into data topology and its interactive 
	exploration by direct manipulation on a whole or on a fragment 
	of dataset. 
For example, one can remove irrelevant data samples, identify the outliers 
	or zoom-in a particular fragment of data. By changing visualization modes
	(e.g. the form of the cost function),
	multi-scale structure of the data can be investigated. 
Summarizing, interactive visualization allows for:

\begin{enumerate}
\item formulation and instant verification of a number of
	hypotheses;
\item precise matching of data mining tools to the properties of data
	investigated;
\item optimize their parameters.
\end{enumerate}

As the famous Visual Analytics Mantra by Keim et al.~\cite{keim2010visual} says:
	{\it `analyze first, show the important, zoom, filter and analyze
	further, details on demand'}. Herein, we focus on application of
	data embedding (DE) methods~\cite{yang2005data} in interactive visualization of
	large ($M\sim 10^{5+}$) high-dimensional datasets.

Data embedding (or shortly, embedding) is defined as a transformation \mbox{$\boldsymbol{B}\!:\mb{Y}\!\rightarrow\!\mb{X}$}
	of $N$-dimensional ($N$-D) dataset 
	$\varmathbb{R}^N\!\ni\!\mb{Y}=\{\mb{y}_i\!=\!(y_{i1},\dots y_{iN})\}_{i=1,\dots M}$
	into its $n$-dimensional `shadow' 
	$ R^n\!\ni\!\mb{X}=\{\mb{x}_i\!=\!(x_{i1},\dots$ $x_{in})\}_{i=1,\dots M}$,
	where $n\ll N$ and $M$ is the number of samples. 
The $\boldsymbol{B}$ mapping should be understood as a lossy compression 
	of $N$-dimensional dataset $\mb{Y}$ in $n$-dimensional space.
The mapping is realized by minimization of a cost function $E(\|\mb{Y}-\mb{X}\|)$, 
	where $\|.\|$ is a measure of topological dissimilarity between
	$\mb{Y}$ and $\mb{X}$. 
Particularly, $E(.)$ should be sensitive to the  multiscale cluster structure of $\mb{Y}$. 
Due to the complexity of low-dimensional manifold -- immersed in $N$-D feature
	space -- on which data samples Y are located, perfect embedding of
	$\mb{Y}$ in $n$ dimensions, with 0 compression error, is possible only for trivial
	cases. 
In fact, the embedding task is a simplification of a more generic problem, where $\boldsymbol{B}$ 
	is a mapping of a given structure $\mb{Y}$ from a
	source space into a corresponding structure $\mb{X}$ in a target space,
	assuming the smallest reconstruction error.

In the context of $N$-dimensional data embedding in visually perceptible
	Euclidean space, we assume here that
	$n=2$, having in mind that 3-D data visualization can be realized in
	a similar way. 
As shown in a large body of papers (see, e.g.,~\cite{yang2005data,van2009dimensionality,france2011two}), 
	DV of large high-dimensional datasets ($M\sim 10^{5+}$ and $N\sim 10^{3+}$)
	that is both sufficiently precise in reconstruction of $N$-D data
	topology and computationally affordable, is an algorithmic
	challenge. 
To preserve topological properties of $\mb{Y}$ in $\mb{X}$ both the
	classical MDS (multidimensional scaling) methods and the state-of-the-art clones of the 
	stochastic neighbor embedding (SNE) concept~\cite{hinton2003stochastic}, 
	require computiing and storing of all the distances
	between data samples both for the source $\mb{Y}$ dataset and its 2-D
	embedding $\mb{X}$. This makes the most precise state-of-the-art vizualization
	algorithms, such as t-SNE~\cite{maaten2008visualizing} and its clones, suffer $O(M^2)$ from time
	and memory complexity. 
The time complexity can be decreased to $O(M\log M)$ by using the approximate 
	versions of t-SNE, such as bh-SNE~\cite{van2014accelerating} and other its 
	variants and simplifications~\cite{yang2014optimization,ingram2015dimensionality,pezzotti2016hierarchical,tang2016visualizing,linderman2017efficient}. 
It is worth to mention here that, in fact, the time-complexity of DE is greater, 
	and depends also on the algorithm which is used for
	the cost function minimization, i.e., on the number of learning
	cycles required to obtain the global minimum of $E(.)$. 
In the rest of this paper -- as it is in all the works about DE -- as the time-complexity 
	we understand here a single learning cycle performed on the full dataset. 
Summing up, as interactive and visual data exploration
	involves very strict time performance regimes, sophisticated data
	structures and high memory load, the state-of-the-art DE
	algorithms can be too slow for visual exploration of data consisting of $M\sim 10^5$ samples.

In this paper we propose a novel and very fast method for
	visualization of high-dimensional data - {\bf i}nteractive {\bf v}isualization
	of {\bf h}igh-dimensional {\bf d}ata technique ({\bf ivhd}) - with linear-time and
	linear-memory complexities. It outperforms the state-of-the-art
	DV algorithms in computational speed retaining high quality of
	data embedding. To this end, we made the following contributions:

\begin{enumerate}
\item We demonstrate that only a small fraction of distances
		(proximities) between $N$-D data samples $\mb{y}_i \in \mb{Y}$ are required --
		just a few nearest and random distances -- to obtain a high
		quality 2-D embedding.
\item Moreover, we show that these distances can be binary, i.e.,
		equal to 0 to the nearest and 1 to the random neighbors of
		each $\mb{y}_i \in \mb{Y}$. 
	Then, instead of using floating point distances, one
		need to find only the indices of a few nearest neighbors ($nn$)
		for each data sample $\mb{y}_i$.\footnote{Here, instead of index $k$
		representing usually the number of the nearest neighbors, we
		use $nn$ to be consistent with the other notation used in this
		paper.} 
	Consequently, the requirements of memory load
		shrinks from $M{\cdot}(M-1)$ floating points (two full distances
	arrays $\mb{D}$ and $\mb{d}$) to barely $nn{\cdot}M$ integers, where $nn\sim 3$.
\item This way we have reduced also the overall time complexity of DE
		method to $O(M)$.
\item We clearly show that data embedding (DE) is equivalent to
		visualization of $nn$-graph built for input $\mb{Y}$ data.
\end{enumerate}

Summarizing, the principal contribution of this paper is the
	essential improvement of time/memory complexity of data
	embedding (with a minor deterioration of data embedding
	quality) what allows for interactive data manipulation. 
The method preserves very well the coarse grained cluster structure of
	the original $N$-D data and enables interactive visualization of
	$M\sim 10^{5+}$ data samples on a laptop. Consequently, our
	improvements open a new perspective to visual exploration of
	much bigger data on more powerful computer systems.

In the following section we demonstrate the details of our
	approach, while in the rest of the paper we discuss its advantages
	and disadvantages in the context of both the state-of-the-art DE
	methods and interactive visualization of truly large datasets. 
In support to our concept we present the results of 2-D embeddings
	of popular benchmark datasets such as small NORB, MNIST,
	twenty newsgroups (20NG) and RCV1. We summarize our
	findings in the Conclusions section.

\section{IVHD data embedding}

Our method represents a radical simplification of the force-directed implementation 
	of the multidimensional scaling (MDS) mapping~\cite{dzwinel1997virtual,pawliczek2013interactive,pawliczek2014visual,pawliczek2015visual}. 
Because it contrasts to the modern SNE concept, exploited in the state-of-the-art DE papers~\cite{hinton2003stochastic,maaten2008visualizing,van2014accelerating,yang2014optimization,ingram2015dimensionality,pezzotti2016hierarchical,tang2016visualizing}, 
	first, we present shortly these two approaches.

\subsection{Background}

We assume that each $N$-D feature vector $\mb{y}_i\in \mb{Y} \subset \varmathbb{R}^N$ is represented
	in 2-D Euclidean space by a corresponding point
	$\mb{x}_i=(x_{i1},x_{i2}) \in \mb{X}\subset \varmathbb{R}^2$ and $i=1,\dots M$. 
We define two distance matrices: $\mb{D}=\{\delta_{ij}\}_{M\times M}$ and 
	$\mb{d}=\{d_{ij}\}_{M\times M}$, in the {\it source} $\varmathbb{R}^N$ and {\it target} $\varmathbb{R}^2$ spaces,
	respectively. 
The value of $\delta_{ij}$ is a measure of dissimilarity (proximity) between feature 
	vectors $\mb{y}_i$ and $\mb{y}_j$ while in 2-D Euclidean space: 
	$d_{ij}=\sqrt{\|\mb{x}_i - \mb{x}_j\|}$. 
In general, the source space has not to be Cartesian one, and can be solely defined by a
	proximity matrix $\mb{D}$ between any two data objects of a (possibly)
	different data representation than the vector one (e.g., shapes,
	graphs etc.).

The classical multidimensional scaling (MDS), which performs
	the mapping $\boldsymbol{B}:\mb{Y}\rightarrow \mb{X}$, consists in minimization of the cost
	(stress) function

\begin{equation}\label{eq:eq1}
E(\|\mb{D}-\mb{d}\|)=\sum_{ij}^M w_{ij}\left(\delta_{ij}^k - d_{ij}^k\right)^m,
\end{equation}

\noindent which represents the error between dissimilarities $\mb{D}$ and
	corresponding distances $\mb{d}$, where: $i,j=1,\dots M$, $w_{ij}$ are weights and $k$,
	$m$ are the parameters. 
Herein, we assume that $m=2$ and $k=1$.
However, there are many other forms of this stress function,
	which are defined, e.g., in~\cite{yang2005data,van2009dimensionality,france2011two}. 
One can easily adopt our approach to particular ($k$, $m$) parameters choice. 
Anyway, finding the global minimum of this multidimensional and multimodal
	cost function (\ref{eq:eq1}) in respect to $\mb{X}$, is not a trivial problem. 
To this end, we use the force-directed approach presented in~\cite{dzwinel1997virtual,pawliczek2013interactive,pawliczek2014visual,pawliczek2015visual}.

We assume that the set of points $\mb{X}$ is treated as an ensemble of
	interacting particles $\mb{x}_i$. The particles evolve in $\varmathbb{R}^2$ space with
	discrete time $n$ scaled by the timestep $\Delta t$, according to the
	Newtonian equations of motion discretized by using the {\it leapfrog} numerical scheme.
Here we use their simplified form~\cite{dzwinel2017ivga}:

\begin{equation}\label{eq:eq6}
\Delta\mb{x}_i \leftarrow a{\cdot}\Delta\mb{x}_i+b{\cdot}\mb{f}_i,
\end{equation}

\begin{equation}\label{eq:eq4}
\mb{f}_i^n = -\nabla\left(\sum_{j=1}^M\left(\delta_{ij}^n - d_{ij}^n\right)^2  \right),
\end{equation}

\begin{equation}\label{eq:eq7}
\mb{x}_i \leftarrow \mb{x}_i + \Delta\mb{x}_i,
\end{equation}

\noindent what resembles well known momentum minimization method. 
The value of $a\in [0,1]$ represents friction, and it is equal to 1 in the absence
	of energy dissipation. 
Meanwhile, $b$ parameterize interparticle forces. 
The proper balance of $a$ and $b$ is crucial for the particle
	system convergence speed to a stable and good (close to the
	global one) minimum of the stress function (\ref{eq:eq1}). 
We demonstrated in~\cite{dzwinel1997virtual,pawliczek2013interactive,pawliczek2014visual,pawliczek2015visual} 
	that this formulation of the classical MDS algorithm
	produces acceptable embeddings for low-dimensional ($N<10$)
	and rather small datasets ($M\sim 10^3$) in sublinear time complexity,
	i.e., similar to widely used stochastic gradient descent (SGD)
	algorithms and its clones employed, e.g., in the original t-SNE
	implementation~\cite{maaten2008visualizing}. 
The principal problems with MDS for $N>10$ are both the
	`curse of dimensionality' effect, which produces bad quality
	embeddings, and high computational and storage complexity for
	$M\sim 10^{5+}$.

As shown in~\cite{beyer1999nearest}, under the broad set of conditions, as
	dimensionality increases, the distance to the nearest feature
	vector approaches the distance to the furthest one for as few as
	10-15 dimensions. 
Consequently, this low distance contrast causes that MDS completely 
	fails in mapping of highdimensional data to 2-D space, producing 
	meaningless sphere of 2-D points for most of high-dimensional datasets. 
As shown in~\cite{hinton2003stochastic,maaten2008visualizing}
	this adverse effect can be radically mitigated by change of
	the definition of proximity measure and the cost function. 
The group of methods based on stochastic neighbor embedding, like
	the most popular t-SNE~\cite{maaten2008visualizing}, defines the similarity of two samples
	$i$ and $j$ (both in the {\it source} and {\it target} spaces) in terms of
	probabilities ($p_{ij}$ and $q_{ij}$, respectively) that i would pick j as its
	neighbor and vice versa. Then the cost function $E(\mb{D},\mb{d})$ is defined
	as the K-L divergence:

\begin{equation}\label{eq:eq8}
E(\mb{D},\mb{d})=
\sum_i\sum_j p_{ij}\log\frac{p_{ij}}{q_{ij}},
\end{equation}

\noindent where, for t-SNE algorithm, $p_{ij}$ is approximated by a Gaussian
	with variance $\sigma$ centered in $\mb{y}_i$, while $q_{ij}$ is defined by the heavy-tailed Cauchy 
	distribution in $\mb{X}$, to avoid both the crowding and the
	optimization problems of older SNE method. 
The respective probabilities are defined as follows:

\begin{equation}\label{eq:eq9}
p_{ij}=\frac{\exp\left(-\frac{\delta_{ij}^2}{2\sigma^2}\right)}
{\sum\limits_{k\neq l}\exp\left(-\frac{\delta_{kl}^2}{2\sigma^2}\right)}
\quad\text{and}\quad
q_{ij} = \frac{\left(1+d_{ij}^2\right)^{-1}}{\sum\limits_{k\neq l}\left(1+d_{kl}^2\right)^{-1}}.
\end{equation}

As demonstrated in~\cite{maaten2008visualizing}, t-SNE outperforms other data reduction
	techniques producing very precise embeddings for all kind of
	datasets, including high-dimensional ones. 
However, unlike in the classical MDS presented above, we cannot use in t-SNE (and
	its clones) the force-directed approach for $E(.)$ minimization due
	to partition function in the denominator in Eqs.~(\ref{eq:eq9}). 
The gradient descent (GD) methods (such as stochastic GD or asynchronous
	stochastic) more often stuck in local minima, what causes that the
	quality of embedding is very sensitive on the choice of
	parameters~\cite{tang2016visualizing}. 
Moreover, similar to MDS, at least $n_{\max}=M{\cdot} (M-1)/2$
	distances $\delta_{ij}$ in $\mb{Y}$, and the same number of corresponding
	distances $d_{ij}$ in $\mb{X}$, have to be calculated. While $\mb{D}$ is computed
	only once at the beginning of simulation, $\mb{d}$ matrix has to be
	computed in every timestep. Moreover, the two distance matrices
	has to be kept in the operational memory. 
This results in prohibitively high $O(M^2)$ time complexity (in respect to
	a single training cycle of minimization method applied) and
	storage. 
This excludes the two methods as acceptable candidates
	for embedding engine of interactive visualization of large
	datasets. 
Moreover, in terms of parallel computations on multicore CPUs, the maximal 
	number of samples $M$ which can be embedded can scale, at most, 
	with the square root of the number of CPU (or GPU) threads involved 
	in computations~\cite{pawliczek2014visual,pawliczek2015visual}.
Anyway, the quadratic memory load remains the key problem
	and puts the upper limit on $M$. The application of Burnes-Hut
	approximation in both MDS and t-SNE (bh-SNE) algorithms~\cite{van2014accelerating}
	can reduce their computational complexity to $O(M\log M)$ but at
	the cost of increase of algorithmic complexity and, consequently,
	decrease of parallelization efficiency. Despite of reduction of the
	number of distances computations in bh-SNE, the method still
	requires full distances table with $M{\cdot} (M-1)$ floating points. 
Do we really need so many distances to embed $N$-D data in 2-D Euclidean
	space?

\subsection{Key concept}

The linear-time and memory complexity of data embedding can
	be achieved by using only a limited number of distances from $\mb{D}$
	and corresponding distances from $\mb{d}$ (such as in~\cite{dzwinel1997virtual,pawliczek2013interactive,pawliczek2014visual,pawliczek2015visual,ingram2009glimmer}). 
In the context of, so called, theory of structural rigidity~\cite{thorpe1999rigidity}, all
	distances between the samples from a $n$-D dataset are not needed
	to retain rigidity of its shape in a $n$-D space. 
The term `rigidity' can be understood as a property of a $n$-D structure made of rods
	(distances) and joints (data vectors) that it does not bend or flex
	under an applied force. 
Therefore, to ensure the rigidity of $\mb{Y}$ (and
	its 2-D embedding $\mb{X}$) only a fraction of distances (joints) from $\mb{D}$
	(and~$\mb{d}$) is required. What is a minimal number of distances for
	which both the original $\mb{Y}$ and target $\mb{X}$ datasets remain rigid?

As shown in~\cite{thorpe1999rigidity,borcea2004number}, minimal $n$-rigid structure, which consists
	of $M\geq n$ joints (vectors) in $n$-dimensional space, requires at least:

\begin{equation}\label{eq:eq10}
L(n)_{\min}=n{\cdot} M - \frac{n{\cdot}(n+1)}{2}
\end{equation}

\noindent rigid rods (distances). It means that, in the source space, the
	number of distances $L(N)_{\min}$ can ensure lossless reconstruction of
	data structure in $N$-D. 
Particularly, in 2-D the structural rigidity can
	be preserved for $L(2)_{\min}=2{\cdot} M-3$. However, $L_{\min}$ defines only the
	lowest bound of $L$ required to retain structural rigidity.
Therefore, herein we are trying to answer the following questions:

\begin{enumerate}
\item What minimal number of distances in $\mb{Y}$ should be known to
		obtain the reasonable reconstruction of data in $\mb{X}$,
		simultaneously, preserving structural rigidity of $\mb{X}$? 
	Is it closer to $N$ or rather to 2?
\item Which distances should be retained?
\end{enumerate}

\begin{figure}
\begin{center}
\includegraphics[width=12cm]{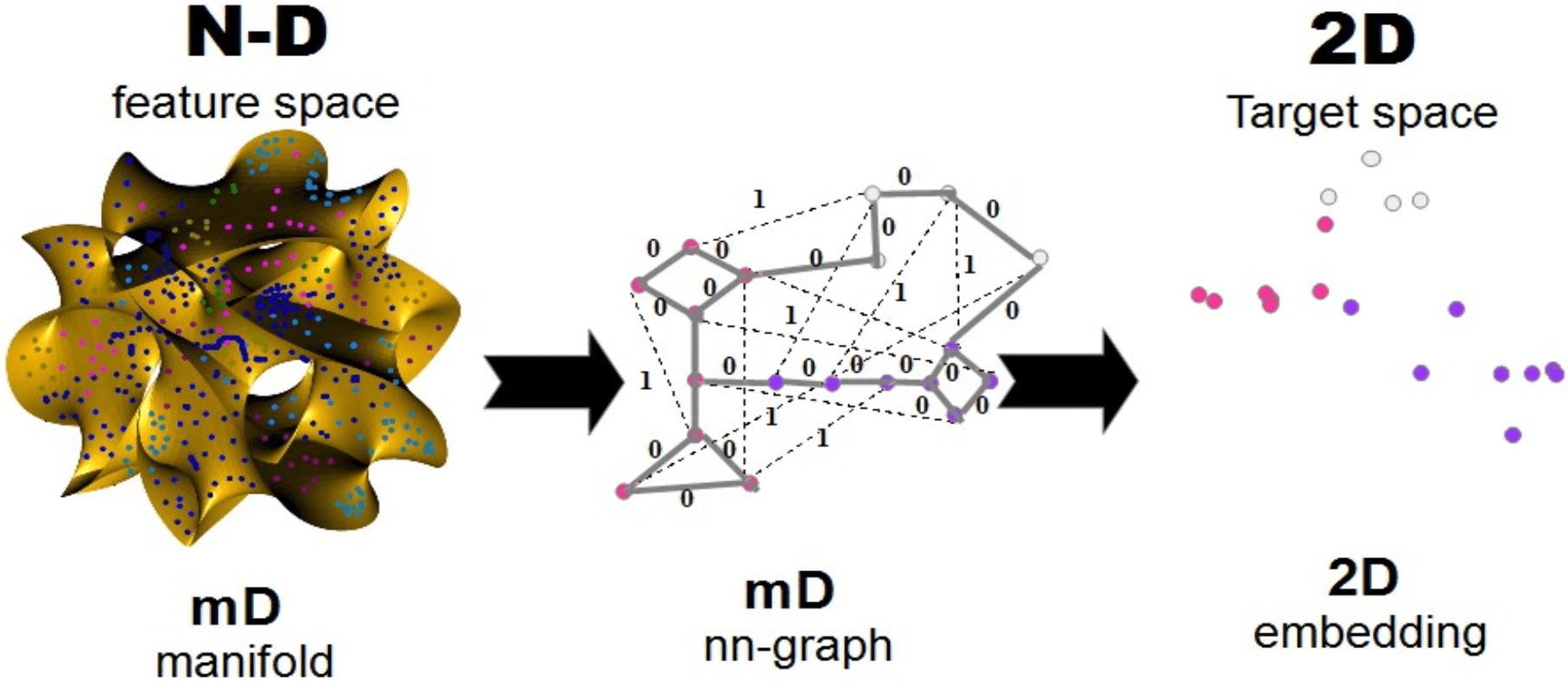}
\end{center}
\caption{The idea of data embedding by means of the nearest
neighbor graph.}
\label{fig:fig1}
\end{figure}

\begin{figure}
\begin{center}
\includegraphics[width=12cm]{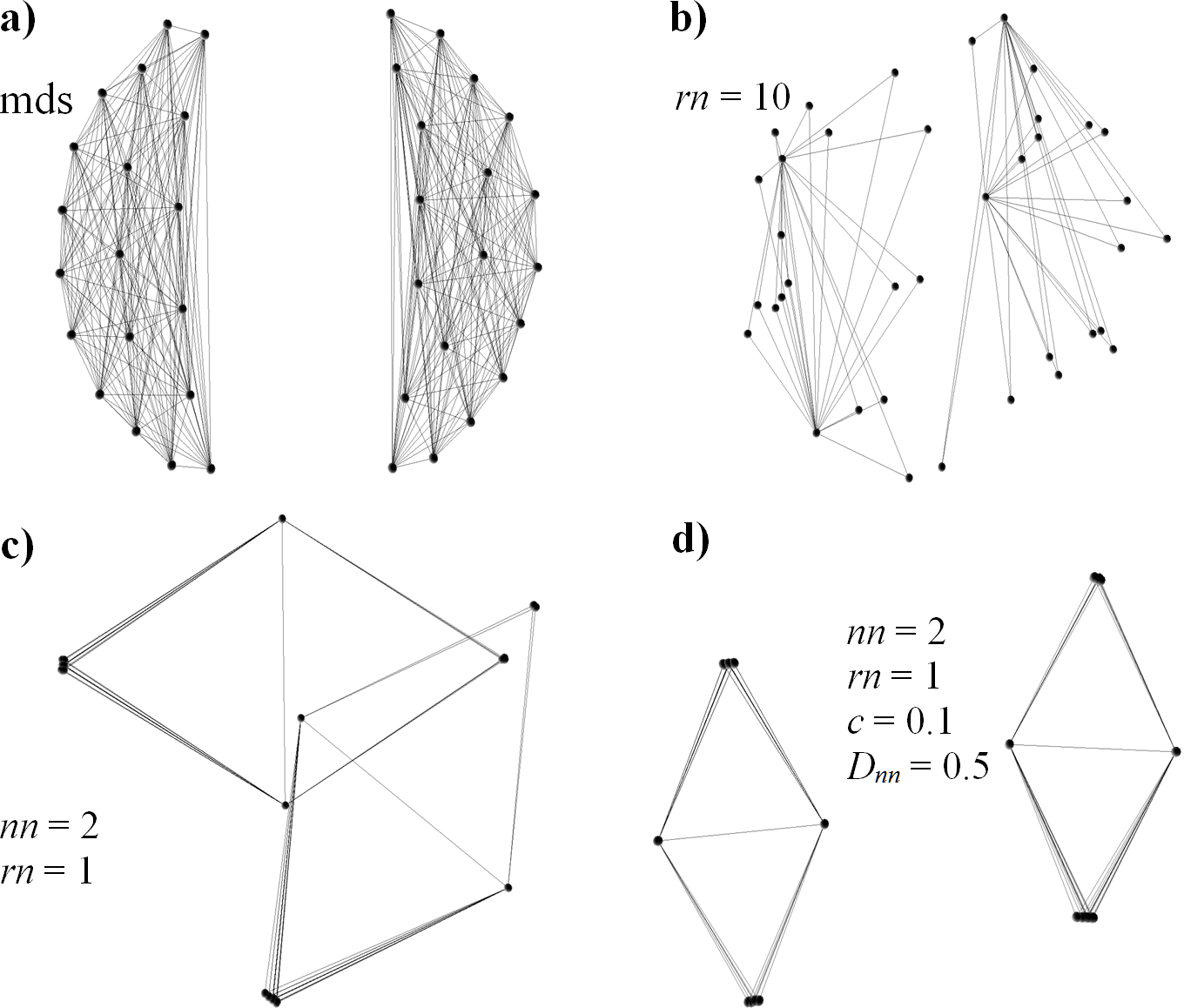}
\end{center}
\caption{The results of 2-D embedding of two identical
regular hypertetrahedrons for (a) original MDS setting and
(b-d) highly reduced number of distances in the cost function
(1).}
\label{fig:fig2}
\end{figure}

In general, $\mb{D}$ cannot be the Euclidean matrix. It may represent
	proximity of samples in an abstract space. 
Particularly, the samples $\mb{y}_i\in \mb{Y}$ can occupy a complicated 
	$n$-D manifold for which $n\ll N$ (see Figure~\ref{fig:fig1}). 
Then, one can assume, that only the distances
	of each sample $\mb{y}_i$ to their $nn$ nearest neighbors are Euclidean. 
As shown in~\cite{tenenbaum2000global,dzwinel2017ivga}, the $nn$ 
	nearest neighbor graph ($nn$-graph), in which vertices represent data 
	samples while edges the connections to their $nn$ nearest neighbors, 
	can be treated as an approximation of this low-dimensional manifold, 
	immersed in a high-dimensional feature space (see Figure~\ref{fig:fig2}). 
In CDA and Isomap~\cite{yang2005data,tenenbaum2000global} DE algorithms, 
	the proximities of more distant graph vertices (samples) are calculated as 
	the lengths of the shortest paths between them in a respective $nn$-graph. 
However, though this way we can more precisely approximate the real
	distances on the low-dimensional manifold, the full distances
	matrix $\mb{D}$ has to be calculated by very time consuming $O(M^3)$
	Dijkstra's (or Floyd-Warshall~\cite{floyd1962algorithm}) algorithm and, what is even more demanding,
	has to be stored in the operational memory. 
In fact, the calculation of full $\mb{D}$ is not necessary, and the problem of data
	embedding can be replaced by the congruent problem of the
	nearest neighbor graph ($nn$-graph) visualization.

The methodology of graph visualization (GV) and DE shares
	many common concepts. For example, the metaphor of particle
	system and the force directed method, used for the cost function
	minimization, were introduced independently to GV and DE~\cite{dzwinel1997virtual,fruchterman1991graph}. 
Summing up, as shown in Figure~\ref{fig:fig1}, the first step in our
	DE method ({\bf ivhd}) consists in construction of $nn$-graph, which
	approximate $n$-dimensional non-Cartesian manifold immersed in $\varmathbb{R}^N$. 
Then we can use the fast procedure of graph visualization in 2-D space
	presented in our recent work~\cite{floyd1962algorithm}. In the rest of this paper we
	demonstrate that the location of graph vertices in 2-D space will
	represent good quality $\mb{Y}$ embedding.

Let us assume that $nnG(\mb{V},\mb{E})$ is the nearest neighbor $nn$-graph
	constructed for high-dimensional dataset $\mb{Y}$. 
The data samples \mbox{$\mb{y}_i\in\mb{Y}$} correspond to the graph vertices 
	$v_i\in \mb{V}$, while $\mb{E}$ is the set of
	edges connecting each $\mb{y}_i$ with their $nn$ nearest neighbors.
Because, we have assumed (see Figure~\ref{fig:fig1}) that this graph is an
	approximation of $n$-dimensional manifold, consequently, we
	presume that its topology preserving embedding in 2-D Euclidean
	space will produce similar results as DE algorithms.
According to definition formulated in~\cite{shaw2009structure} that 
	{\it `topology is preserved if a connectivity algorithm, such as k-nearest neighbors, can easily
	recover the edges of the input graph from only the coordinates of
	the nodes after embedding'}. It means that unlike in DE
	algorithms, which tend to retain the neighbors order, the ordering
	of $nn$ nearest neighbors (usually a few ones) is irrelevant in that
	case. 
In fact, the requirement of preserving the order of small
	number of $nn$ nearest neighbors for each $\mb{y}_i\in \mb{Y}$, which are subject
	to uncertainty and measurement errors, is secondary in terms of
	large data visualization. 
The crucial problem for formulating {\it ad hoc} hypotheses and to 
	decide about the use of particular machine
	learning tools in further data analytics is revealing data topology,
	i.e., its multi-scale cluster structure.

Because we are not interested in $nn$ nearest neighbors
	ordering for each $\mb{y}_i\in \mb{Y}$, to get a good approximation of data
	manifold, to the sake of simplicity, the distances between
	connected vertices of $nn$-graph should be the same and as close to
	0 as possible. 
Conversely, the unconnected vertices of $nnG(\mb{V},\mb{E})$
	has to be more distant, to increase the contrast between both
	types of distances. 
It is obvious that $nn$-graphs (for small $nn$) are
	largely unconnected and not rigid. Therefore, $nnG(\mb{V},\mb{E})$ graph
	has to be augmented by additional edges. 
Here, we assume that for each $v_i\in \mb{V}$ ($\mb{y}_i\in \mb{Y}$) apart 
	from $nn$ edges to the nearest
	neighbors, additional $rn$ randomly selected neighbors are
	connected. 
The value of $rn$ should ensure rigidity of this augmented graph. 
Thus, $O_{\text{\it nn}}(i)$ and $O_{\text{\it rn}}(i)$ are the sets of indices of
	$nn$ nearest (connected) neighbors and $rn$ (unconnected) random
	neighbors of a feature vector $i$ ($\mb{y}_i$) in $\mb{Y}$, respectively. 
We assume also that $O_{\text{\it nn}}(i) \cap O_{\text{\it rn}}(i)=\emptyset$. 
However, this assumption is superfluous because the probability of picking the nearest
	neighbor as a random neighbor is negligible low for large $M$. 
In fact, the random neighbors are very distant from $\mb{y}_i$. 
Therefore, having in mind the effect of the `curse of dimensionality',
	we can radically simplify data
	similarity to the binary one \{`nearest'~(0), `random'~(1)\}, i.e.:

\begin{equation}\label{eq:eq11}
\forall \mb{y}_i \in \mb{Y}:
\left\{
\begin{array}{lll}
\delta_{ij}=0 & \text{\it iff} & j \in O_{\text{\it nn}}(i) \\
\delta_{ij}=1 & \text{\it iff} & j \in O_{\text{\it rn}}(i) \\
\end{array}
\right.
.
\end{equation}

This way instead of $\mb{D}$ matrix we have as the input data a lists of
	the nearest neighbors of $\mb{y}_i$ ($nn$-graph edges). 
In the rest of this paper, the respective distances from Eq.~(\ref{eq:eq11}) we call shortly 
	$D_{\text{\it nn}}$ and
	$D_{\text{\it rn}}$ (i.e., $D_{\text{\it nn}}=0$ and $D_{\text{\it rn}}=1$).

Summarizing, we can presume that identification of a few $nn$
	nearest neighbors and $rn$ random neighbors for each $\mb{y}_i\in \mb{Y}$, where
	$n_{vi}=nn+rn$ is small, is sufficient to assure rigidity of augmented
	$nnG(\mb{V},\mb{E})$ graph. 
This is rather obvious, because it is well known~\cite{cohen1997three}
	that any graph can be embedded preserving its topology (in the
	context of definition presented above), at most, in 3-D Euclidean
	space while planar graphs in 2-D. 
In the most cases $nn$-graphs for small $M$ and $N$ are
	planar, though for $nn>1$ the probability of non-planarity rapidly
	increases with $M$~\cite{wilkinson2005graph}. 
Thus the total number of connections $L =
	|\mb{E}|$ required for making the augmented $nn$-graph rigid is of order
	$L\sim n_{vi}{\cdot} M$. 
In fact, we need to store only $nn$ indices of the nearest
	neighbors for every $\mb{y}_i\in \mb{Y}$, while the indices of $rn$ random
	neighbors can be generated {\it ad hoc} during embedding process.

Additionally, we presume that by imposing a high contrast on
	these two types of distances, we will be able to preserve in $\mb{X}$ the
	multi-scale cluster structure of $\mb{Y}$, by employing MDS mapping
	only on those $L$ binary distances. 
This way we could decrease both the
	computational complexity of data embedding to $O(a{\cdot} M)$ with
	\mbox{$a\sim n_{vi}$} and its computational load to only $nn{\cdot} M$ integers.

To obtain 2-D embedding $\mb{X}$ of $nnG(\mb{V},\mb{E})$ we minimize the
	following stress function:

\begin{equation}\label{eq:eq12}
E(\|\mb{D}-\mb{d}\|)=
\sum_i
\ \sum_{j\in O_{\text{\it nn}}(i)\cup O_{\text{\it rn}}(i)}
\!\!\!\!\!\!\!\!\!\!\!b\left(\delta_{ij}-d_{ij}\right)^2.
\end{equation}

Consequently, the interparticle force $\mb{f}_i$ from Eq.~(\ref{eq:eq4}) in $E(.)$
	minimization procedure simplifies to:

\begin{equation}\label{eq:eq13}
\mb{f}_i^n=
-\!\!\!\!\!\sum_{j\in O_{\text{\it nn}}(i)}^{nn}\!\!\!\!\!\mb{x}_{ij}^n
-c\!\!\!\!\!\sum_{k\in O_{\text{\it rn}}(i)}^{rn}
\!\!\left(1-d_{ik}^n\right){\cdot}\ \frac{\mb{x}_{ik}^n}{d_{ik}^n},
\quad \mb{x}_{ik}^n = \mb{x}_i^n-\mb{x}_k^n.
\end{equation}

To explain our concept in terms of data embedding, let us follow
	a toy example.

We assume that $\mb{Y}$ consists of $M=38$ samples located in the
	vertices of two identical but translated and mutually rotated,
	regular 18-dimensional hypertetrahedrons. In fact, data create an
	unconnected graph, which vertices are joined by the
	hypertetrahedron edges. 
As shown in Figure~\ref{fig:fig2}a, by employing
	classical MDS with the cost function (\ref{eq:eq1}) and the dataset defined
	by the full distances table $\mb{D}$ ($L=703$ distances), we have obtained
	$\mb{X}$ embedding shown in Figure~\ref{fig:fig2}a. 
One can observe that the result is rather not satisfactory. 
Though it demonstrates well data separability, the local structure 
	of $\mb{X}$ remains very unclear.
Let us assume now that we know only $rn=10$ distances from $\mb{D}$ to
	random neighbors for every $\mb{y}_i\in \mb{Y}$ ($L=300$ distances). 
As one can expect, the final embedding from Figure~\ref{fig:fig2}b shows that data
	separation is still visible but it lacks any fine grained structure.
As shown in Figure~\ref{fig:fig2}c, this flaw can be partially eliminated
	taking into account also distances to $nn$ nearest neighbors of $\mb{y}_i$,
	i.e., in this particular case $nn=2$ and $rn=1$ were set ($L=103$
	distances). 
Additionally, we have reduced the strong bias caused
	by the long-range interactions between $\mb{y}_i$ samples and their
	random neighbors, by decreasing 10 times the
	scaling factor of the respective `forces' ($c=0.1$ in Eq.~(\ref{eq:eq13})). 
	Unlike in Figure~\ref{fig:fig2}b, in
	Figure~\ref{fig:fig2}c we observe much better reconstruction of local $\mb{Y}$
	structure but at the cost of worse data separation. 
The embeddings of two regular hypertetrahedrons overlap each other.

Let us assume now that all the distances between all $M$ samples
	are equal to 1. To increase the contrast between the $nn$ nearest
	and $rn$ random neighbors of $\mb{y}_i$, we assume that $D_{\text{\it nn}}=0.5$ and $D_{\text{\it rn}}=1$. 
Because now we use binary distances \{0.5, 1\}, we do not need to store floating
	point matrix $\mb{D}$ but just remember $nn$ indices (integers) of each $\mb{y}_i$
	to their nearest neighbors (here 70 integers in total). 
As shown in Figure~\ref{fig:fig2}d, the 2-D embedding clearly reconstructs the local and
	global structure of $nn$-graph representing original 18-dimensional data. 
Because $M$ and $N$ were small in this example the value of
	$D_{\text{\it nn}}=0.5$, was sufficient to properly contrast $nn$ and $rn$ distances.
However, as we show above (see Eq.~(\ref{eq:eq11})), for larger $M$ and $N$, $D_{\text{\it nn}}$
	should be equal to 0 to reduce the `curse of dimensionality' effect. 
It is worth to mention here that the neighborhood relation
	is not symmetrical, thus $(n=nn+rn){\cdot} M$ can be smaller than $L(2)_{\min}$
	for $n=2$. 
Consequently, it may produce non-rigid and deformed (2-D) embeddings. 
Moreover, by assuming that $nn=1$, one can get
	meaningless 2-D embeddings even though $rn$ is much greater than 2.
Summarizing, we come to the following conjectures.

\begin{enumerate}
\item The number of distances from $\mb{D}$, sufficient to obtain 2-D
		embedding $\mb{X}$ of original $N$-D dataset $\mb{Y}$, which preserves both
		local and global properties of $\mb{Y}$, can be surprisingly small
		and close to $\sim 3{\cdot} M$.
\item For each $\mb{y}_i\in \mb{Y}$, their vicinity consisting of a few $nn$ nearest
		neighbors should be preserved while the global cluster
		structure is controlled by a few (often one) $rn$ randomly
		selected neighbors.
\item The distances between $\mb{y}_i\in \mb{Y}$ and their nearest and random
		neighbors, and respective forces in optimization procedure
		should be properly contrasted. For high dimensional data,
		binary distance can be used. This can radically reduce the
		memory load from $M{\cdot} (M-1)$ floating points (two distances
		arrays $\mb{D}$ and $\mb{d}$) to $nn{\cdot} M$ integers, where $nn\sim 3$.
\end{enumerate}

In the following section we will demonstrate that these simple
	improvements can make the classical MDS competitive, both in
	terms of efficiency and embedding quality, competitive to the
	state-of-the-art DE algorithms. 
Particularly, for the interactive visualization of large datasets.

\section{Implementation and tests}

On the basis of DE algorithm described above, we have
	developed an integrated tool for {\bf i}nteractive {\bf v}isualization of 
	{\bf h}igh-dimensional {\bf d}ata ({\bf ivhd}). 
The code was written in C++, while
	GUI was created by using OpenGL\texttrademark{} graphics system integrated
	with the standard Qt GUI interface. 
All the computations and
	visualizations were performed on a notebook computer with
	Intel\textregistered{} Core\texttrademark{} i5 7200U 2.5-2.71GHz with 16GB 
	of memory and equipped with Nvidia GeForce 940MX 2GB graphic board.

The DE initialization, which computes the nearest neighbors
	for the source $\mb{Y}$ dataset, was written in CUDA\texttrademark{} and uses two
	very fast GPU implementations of brute force and approximated
	(based on LargeVis idea~\cite{tang2016visualizing}) algorithms for the nearest
	neighbors search~\cite{klusek2017multi}. 
This search is performed only once at the
	beginning of simulation and, as shown in Table~\ref{tab:tab1}, it requires
	relatively long computational time for MNIST data set with
	$M\sim 10^5$, $N\sim 10^3$ and $nn=100$. Pay attention that GPU
	implementation of brute-force algorithm is faster for larger $nn$
	and $N$ than LargeVis approximated algorithm (see~\cite{klusek2017multi} for
	details). 
The CPU version of these algorithms (MPI, two cores) is
	about 10 times slower. 
The nearest neighbors are cached on the fast SSD disk for further use. 
To evaluate {\bf ivhd} embedding
	concept we have performed experiments on a few highdimensional and multi-class datasets. 
The tests have been carried
	out for four high-dimensional datasets, which are specified in Table~\ref{tab:tab2}.

\begin{table}
\begin{center}
\caption{Selected timings for initialization and $nn$ nearest
neighbors search.}
\label{tab:tab1}
\begin{tabular}{lrr}
\toprule
Data set 
& \multicolumn{1}{l}{Brute-force}
& \multicolumn{1}{l}{Approx.} \\
\midrule
MNIST ($M=7{\cdot} 10^4$, $N\!=\ 30$, $nn\!=\!100$) & 1m 27s & 1m 35s \\
MNIST ($M=7{\cdot} 10^4$, $N\!=\ 30$, $nn\!=\ \ \ 2$) & 1m 17s & 39s \\
MNIST ($M=7{\cdot} 10^4$, $N\!=\!784$, $nn\!=\ \ \ 2$) & 20m 16s & 2m 49s \\
MNIST ($M=7{\cdot} 10^4$, $N\!=\!784$, $nn\!=\!100$) & 20m 29s & 32m 06s \\
\bottomrule
\end{tabular}
\end{center}
\end{table}

\begin{table}
\begin{center}
\caption{The list of data sets referenced in this paper.}
\label{tab:tab2}
\begin{tabular}{lrrrp{8cm}}
\toprule
\multicolumn{1}{l}{Name} 
& \multicolumn{1}{l}{$\ \ N$} 
& \multicolumn{1}{l}{$\ \ M$} 
& \multicolumn{1}{l}{$\,C$} 
& \multicolumn{1}{l}{Short description} \\
\midrule
MNIST & 784 & 70\,000 & 10 & {\small Well balanced set of gray-scale handwritten digit images (http://yann.lecun.com/exdb/mnist).}\\ \noalign{\smallskip}
NORB &\!\!2048 & 43\,600 & 5 & {\small Small NORB dataset (NYU Object
Recognition Benchmark) contains stereo
image pairs of 50 uniform-colored toys
under 18 azimuths, 9 elevations, and 6
lighting conditions
(http://www.cs.nyu.edu/$\sim$ylclab/ data/norbv1.0).} \\ \noalign{\smallskip}
20NG &\!\!2000 & 18\,759 & 20 & {\small A balanced collection of documents from 20
various newsgroups. Each vertex stands for a
text document represented by bag of words
(BOW) feature vector.
(http://qwone.com/$\sim$jason/ 20Newsgroups).} \\ \noalign{\smallskip}
RCV1 & 30 &\!\!804\,409 &  103 & {\small 
Strongly imbalanced text corpus known as RCV1. We transformed an original dataset from https://old.datahub.io/dataset/rcv1-v2-lyrl2004 from $N=2000$ to $N=30$ by using PCA.}\\
\bottomrule
\end{tabular}
\end{center}
\end{table}

Before finding 2-D embeddings we have transformed 20NG and
	Reuters to 30-dimensional space by using PCA transformation.
We have checked that this procedure does not change visibly the
	results of embeddings, but it substantially decreased
	computational time needed in the initialization phase of finding
	the nearest (or the near) neighbors. 
We retained the high-dimensionality for MNIST ($N-784$) and NORB ($N-2048$). 
We used the cosine distance for NORB, Reuters and 20NG while the Euclidean one
	only for MNIST. 
The results obtained by {\bf ivhd} were compared
	with the results generated by two the state-of-art methods: t-SNE
	(its faster bh-SNE version~\cite{maaten2008visualizing}) and its improved and recently
	published clone Large-vis~\cite{pezzotti2016hierarchical}. 
We used the serial codes published in GitHub 
	(http://alexanderfabisch.github.io/t-sne-inscikit-learn.html and https://github.com/lferry007/LargeVis,
	respectively).

Because of extremely high computational load required for
	computing precision/recall coefficients, to compare data
	separability and class purity, we define the following leave one-out coefficients:

\begin{equation}\label{eq:eq14}
cf_{\text{\it nn}}=\frac{\sum_{i=1}^{M}nn(i)}{nn{\cdot} M}
\quad\text{and}\quad
cf=\frac{\sum_{{\text{\it nn}}=1}^{{\text{\it nn}}_{\max}}cf_{\text{\it nn}}}{nn_{\max}},
\end{equation}

\noindent where $nn(i)$ is the number of the nearest neighbors from 
	$O_{\text{\it nn}}(i)$, which belong to the same class as $\mb{y}_i$. 
The value of $cf$ is computed for arbitrary defined value of $nn_{\max}$ 
	dependent on the number of points in classes (here we assume that 
	$nn_{\max}=100$, because the smallest class consists of about 
	1000 samples). 
The value of $cf\sim 1$ for well separated and pure classes, 
	while $cf\sim 1/K$ for random points from $K$ classes.

\begin{figure}
\begin{center}
\includegraphics[width=12cm]{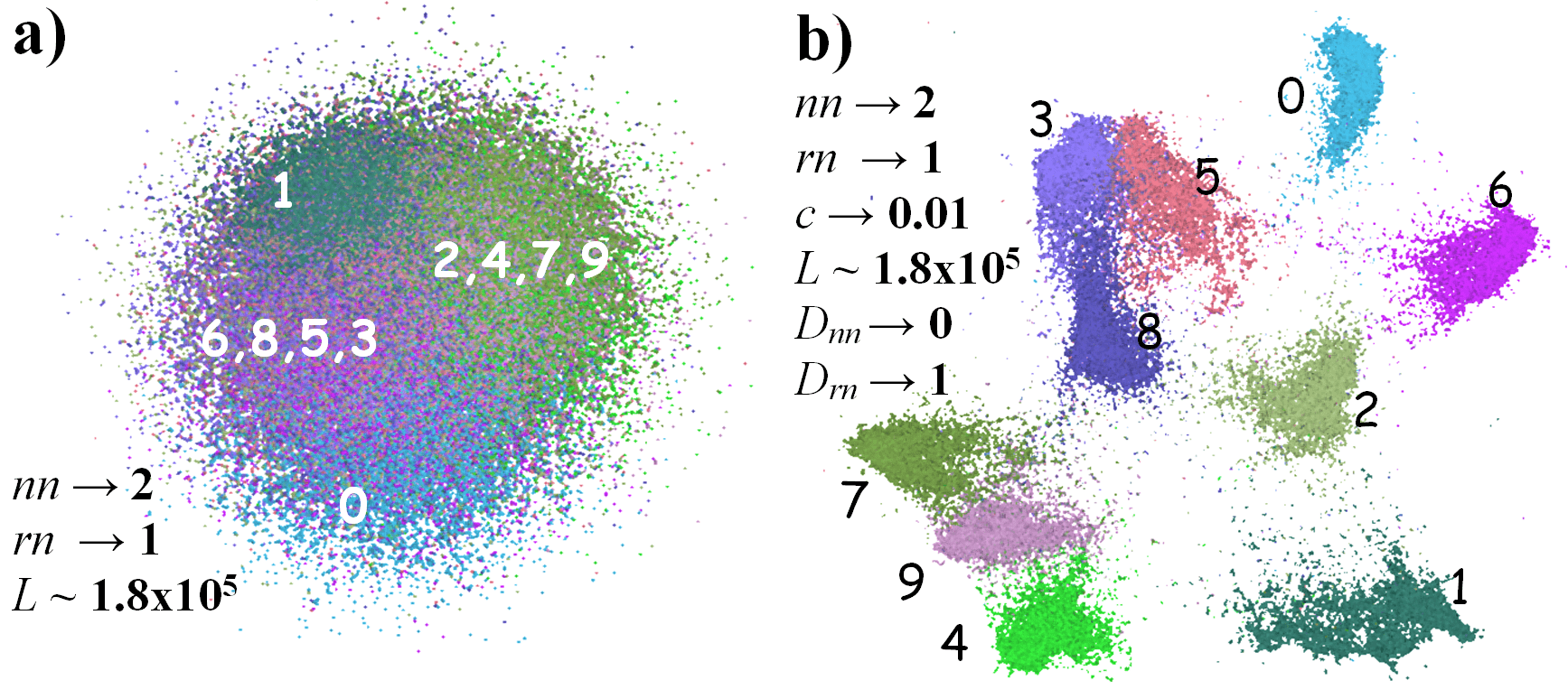}
\end{center}
\caption{The embeddings of MNIST dataset for two
collections of $nn$, $rn$ and $c$ parameters.}
\label{fig:fig3}
\end{figure}

In Figure~\ref{fig:fig3} we demonstrate 2-D embeddings of MNIST dataset
	for two sets of \{$nn$, $rn$, $c$\} parameters. 
In Figure~\ref{fig:fig3}a one can see a typical effect of the `curse of dimensionality'. 
Due to the lack of sufficient contrast of distances the classes are mixed up.
Consequently, we show in Figure~\ref{fig:fig3}b that by assuming binary
	distance between data vectors the configuration of classes is very
	similar to existing visualizations of MNIST (see e.g. L.vd.Maaten
	web page https://lvdmaaten.github.io).

\begin{figure}
\begin{center}
\includegraphics[width=12cm]{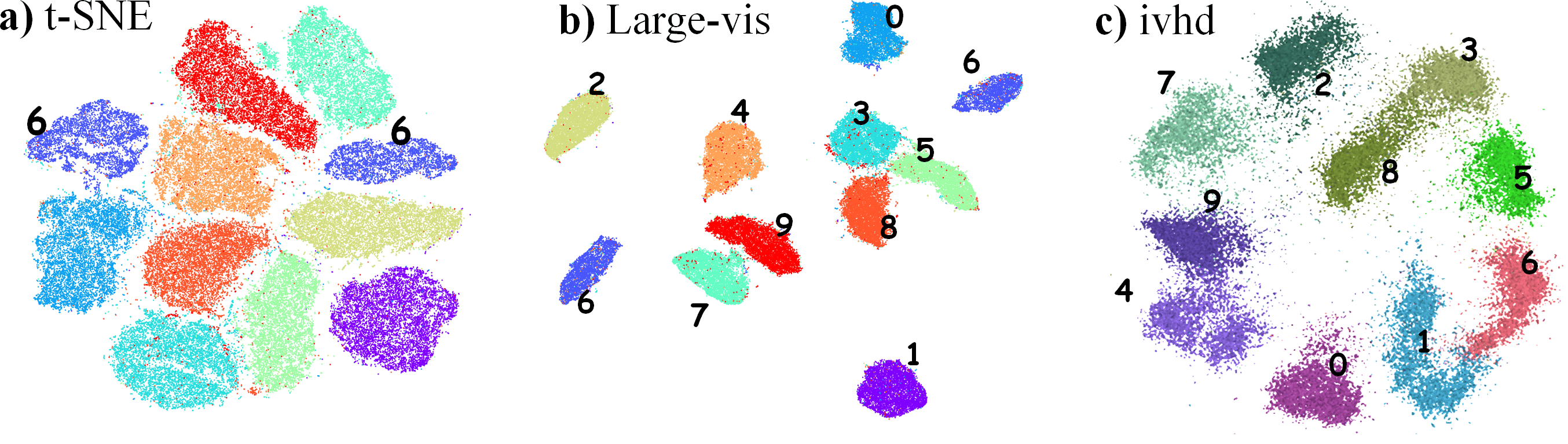}
\end{center}
\caption{The confrontation of MNIST embeddings for three
various algorithms (bh-SNE, Large-vis and {\bf ivhd}). For {\bf ivhd}
we used binary distance and the following lists of parameters:
$nn=2$, $rn=1$ and $c=0.005$. The average values of $cf$ coefficients
are as follows: a) $0.94\pm 0.02$, b) $0.95\pm 0.03$, c)~$0.91\pm 0.02$.}
\label{fig:fig4}
\end{figure}

In Figure~\ref{fig:fig4} we present the confrontation of our method with bhSNE and LargeVis algorithms. 
Unfortunately, we were not able
	to generate ten separate MNIST classes by using these
	competitive methods. 
For the two, the cluster 6 divides into two
	separate classes (similar result can be found in
	http://alexanderfabisch.github.io/t-sne-in-scikit-learn.html).
One can find also the proper 10 class embeddings in the Internet (see
	https://lvdmaaten.github.io/tsne) but there are mainly
	visualizations of the original dataset after PCA transformation to
	30 dimensions. 
It seems that this is the problem of choice of a
	proper parameter set. Moreover, the stochastic gradient descent
	method used for minimization of K-L cost function (Eq.~(\ref{eq:eq8})) can
	stuck in a local minimum, so the minimization procedure should
	be repeated a few times starting from various initial configurations.

On the other hand, comparing visually the quality of embeddings,
	one can see that {\bf ivhd} produces the most blurred solution (it has
	also distinctly lower $cf$ value see Figures~\ref{fig:fig3} and \ref{fig:fig4}). 
However, the coarse grained embeddings are very similar. For example, groups
	of similar clusters \{3, 5, 8\} and \{4, 7, 9\} can be observed both in
	Figure~\ref{fig:fig4}b and \ref{fig:fig4}c. 
Though these groups cannot be found in Figure~\ref{fig:fig4}a, 
	many pictures of MNIST embedding generated by t-SNE
	algorithm, which can be found in the Internet, approve this
	observation. 
Anyway, {\bf ivhd} is unbeatable in terms of computational time. 
The generation of MNIST mapping needs
	from 12 to 40 seconds (this difference depends on the number of
	the nearest and random neighbors used for computations and the
	values of parameters of minimization procedure) while in the
	case of Large-vis and bh-SNE for a single run, without repetition
	for finding a better solution, it is 10 and 20 minutes, respectively.
The approximate timings are given without the computational
	time needed for initialization of the nearest neighbors (see Table~\ref{tab:tab1}). 
However, the high efficiency of {\bf ivhd} is obtained at the
	sacrifice of a quality of reconstruction of the fine grained cluster
	structure and local neighborhood. Similarly to SNE~\cite{hinton2003stochastic}, {\bf ivhd}
	suffers the problem of crowding. The most of samples are located
	in the center of clusters in 2-D embeddings.

\begin{figure}
\begin{center}
\includegraphics[width=12cm]{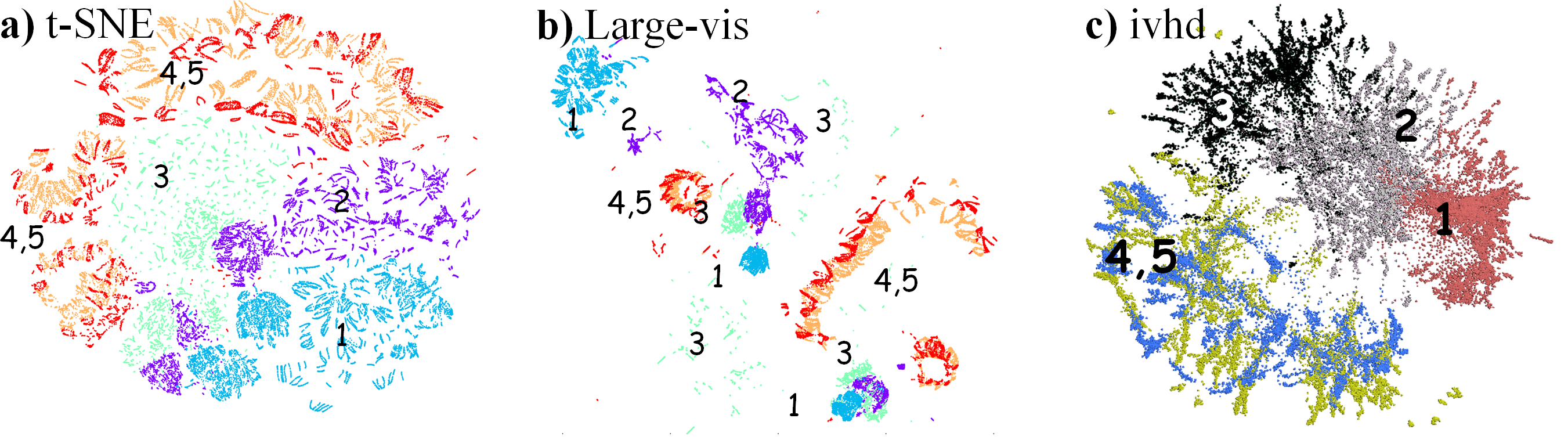}
\end{center}
\caption{The confrontation of NORB embeddings for three
various algorithms (bh-SNE, Large-vis and {\bf ivhd}). For {\bf ivhd}
we used the following lists of parameters: $nn=10$, $rn=1$ and
$c=0.01$. The average values of $cf$ coefficients are as follows: a)
$0.95\pm 0.02$, b) $0.93\pm 0.015$, c) $0.97\pm 0.01$.}
\label{fig:fig5}
\end{figure}

Better local precision of t-SNE and Large-vis visualizations can
	be clearly seen in Figure~\ref{fig:fig5}. Although {\bf ivhd} better separates the
	classes, the sub-clusters representing the same toys in various
	conditions are blurred. 
Meanwhile, in Figures~\ref{fig:fig5}a and \ref{fig:fig5}b one can
	discern the streaks of points representing the same toys. By
	assuming $c=0.01$ in {\bf ivhd}, this local structure of NORB also
	becomes more sharp (the force from the nearest neighbors
	dominates over that from random neighbors), however, at the
	cost of even greater crowding effect. 
Computational time spent by {\bf ivhd} in NORB embedding 30-60 sec. 
	is two times greater than that for MNIST (because $nn=10$), while 
	by using Large-vis and tSNE it grows to about 15 and 20 minutes, respectively.

\begin{figure}
\begin{center}
\includegraphics[width=12cm]{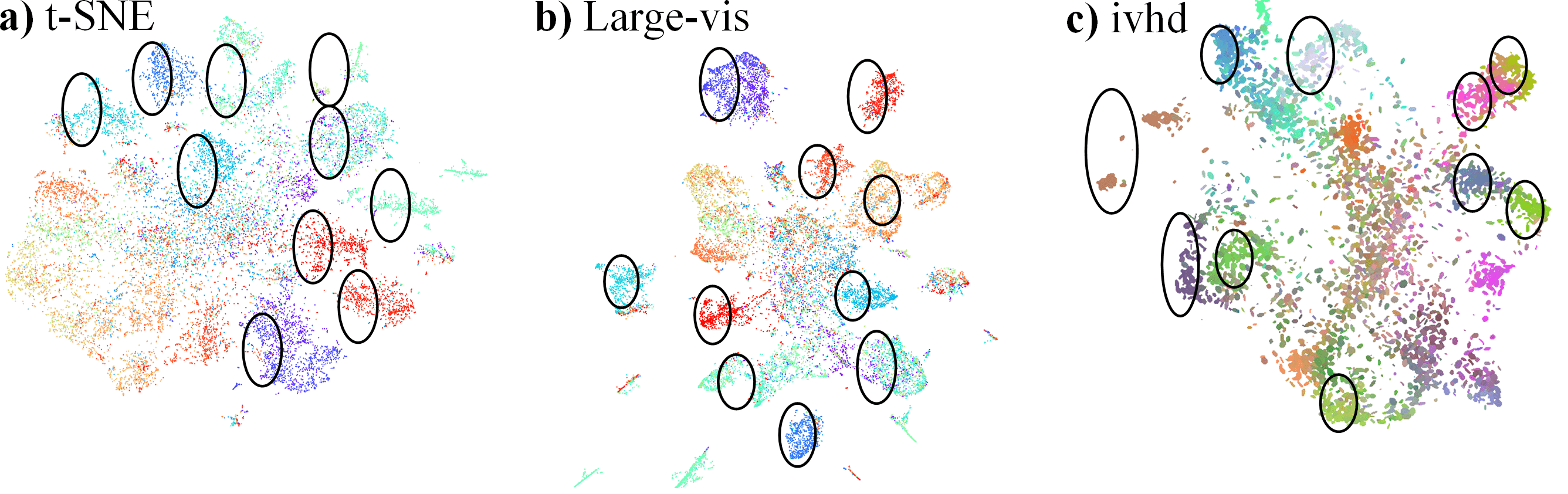}
\end{center}
\caption{The confrontation of 20NG dataset embeddings for
three algorithms (bh-SNE, Large-vis and {\bf ivhd}). For {\bf ivhd} we
used the following lists of parameters: $nn=5$, $rn=3$ and $c=0.1$.
The average values of $cf$ coefficients are as follows: a)
$0.48\pm 0.025$, b) $0.47\pm 0.025$, c) $0.45\pm 0.02$.}
\label{fig:fig6}
\end{figure}

For larger number of overlapped classes and relatively small
	number of data points -- as it is in the twenty newsgroups
	(20NG) dataset -- the embeddings obtained by the three
	algorithms look very messy with relatively low $cf$ coefficients
	(see Figure~\ref{fig:fig6}). In Figure~\ref{fig:fig6} ten most visible clusters are marked
	with circles. 
As it is demonstrated in the web page
	https://lvdmaaten.github.io/tsne, much better cluster separation
	(but of low class purity) can be obtained by using different text
	representation than BOW.

\begin{figure}
\begin{center}
\includegraphics[width=12cm]{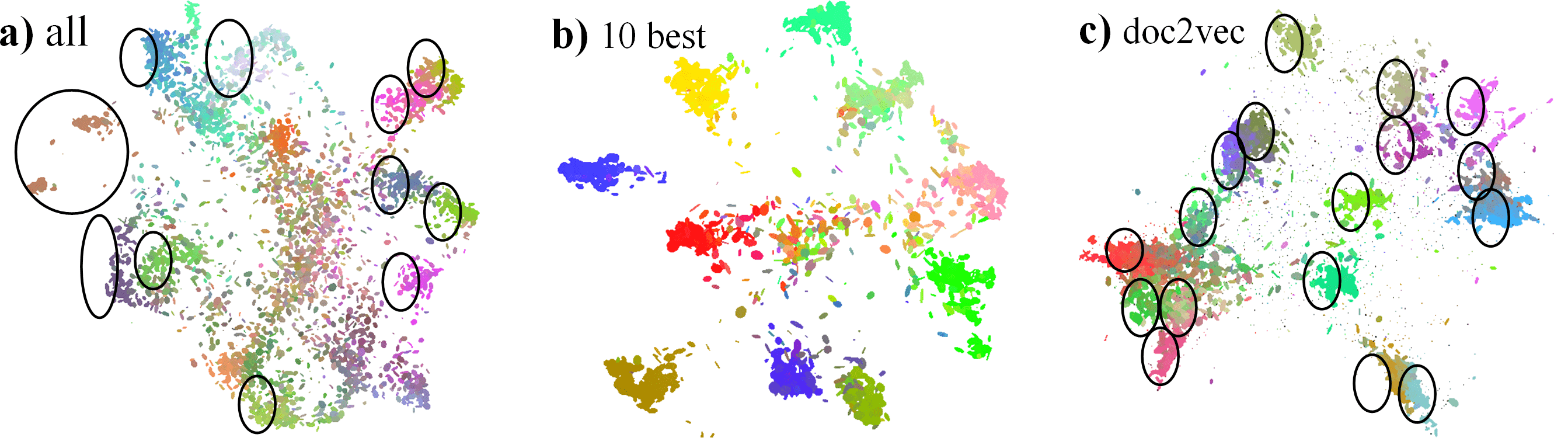}
\end{center}
\caption{The results of {\bf ivhd} embedding of 20NG dataset a)
all clusters, b) ten best separable clusters c) all clusters for
doc2vec representation. The average values of $cf$ coefficients
are as follows: a) $0.43\pm 0.01$, b) $0.73\pm 0.02$, c) $0.52\pm 0.02$.}
\label{fig:fig7}
\end{figure}

\begin{figure}
\begin{center}
\includegraphics[width=12cm]{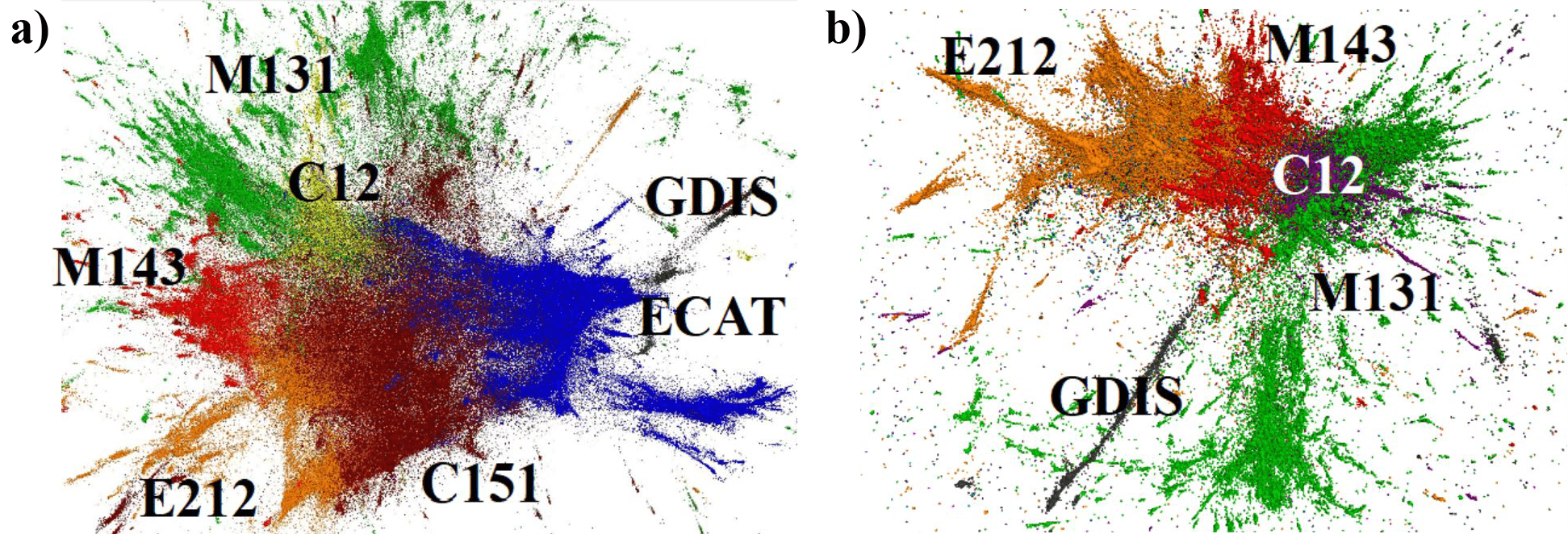}
\end{center}
\caption{The results of RCV1 2-dimensional embedding by using {\bf ivhd}: 
a) All classes visualized ($M\sim8.0\cdot 10^5$) while only the largest 7 are visible; 
b) After removal of the two largest classes (ECAT $\rightarrow$ 224\,149; C151 $\rightarrow$ 185\,582) the 5 classes (E212 $
\rightarrow$ 150\,163; M131 $\rightarrow$ 135\,566; M143 $\rightarrow$ 43\,920; C12 $\rightarrow$ 30\,505; GDIS $
\rightarrow$ 23\,345) are separated better. 
We used the following list of parameters: $nn\!=\!20$, $rn\!=\!3$ and $c\!=\!0.05$. The average 
val\-ues of $cf$ coefficients are as follows: a) $0.82\pm 0.02$, b) $0.93\pm 0.01$.
}
\label{fig:fig8}
\end{figure}

In Figure~\ref{fig:fig7} we compare the {\bf ivhd} embeddings of 20NG datasets
	for BOW and doc2vec~\cite{le2014distributed} representations. 
For the latter, as shown in Figure~\ref{fig:fig7}c, more clusters than those marked in 
	Figure~\ref{fig:fig7}a can be visually recognized. 
This is reflected also by the values of $cf$ coefficient. 
The average computational time required for
	20NG embedding is about 5 sec.
It is more than an order of magnitude lower than those obtained for competitive algorithms.

Unlike in the previous examples, the RCV1 dataset
	consisting of $\sim 8.0{\cdot} 10^5$ samples from scores of partly overlapped 
	and overlapped classes, is highly imbalanced.
Two separate classes ECAT (Economics) and C151 (Accounts/Earnings) represent 64\% of the
	whole dataset. 
This fact is mainly responsible for high $cf$ value
	(see Figure~\ref{fig:fig8}), despite rather poor separation of other categories.
The value of $cf=0.97$ taking into account the 2-D embedding for
	only these two categories. 
However, by excluding them
	from the dataset, we obtain very good embedding (see Figure~\ref{fig:fig8}b)
	of the rest 6 classes with high $cf=0.92$ value. 
We were not able to obtain both t-SNE and Large-Vis embeddings in a reasonable
	time, on the laptop with technical parameters specified at the
	beginning of this section. 
Meanwhile, {\bf ivhd} needs about 7-8
	minutes to produce the result shown in Figure~\ref{fig:fig8}a, and about 5-6
	minutes of that from Figure~\ref{fig:fig8}b.

\begin{figure}
\begin{center}
\includegraphics[width=12cm]{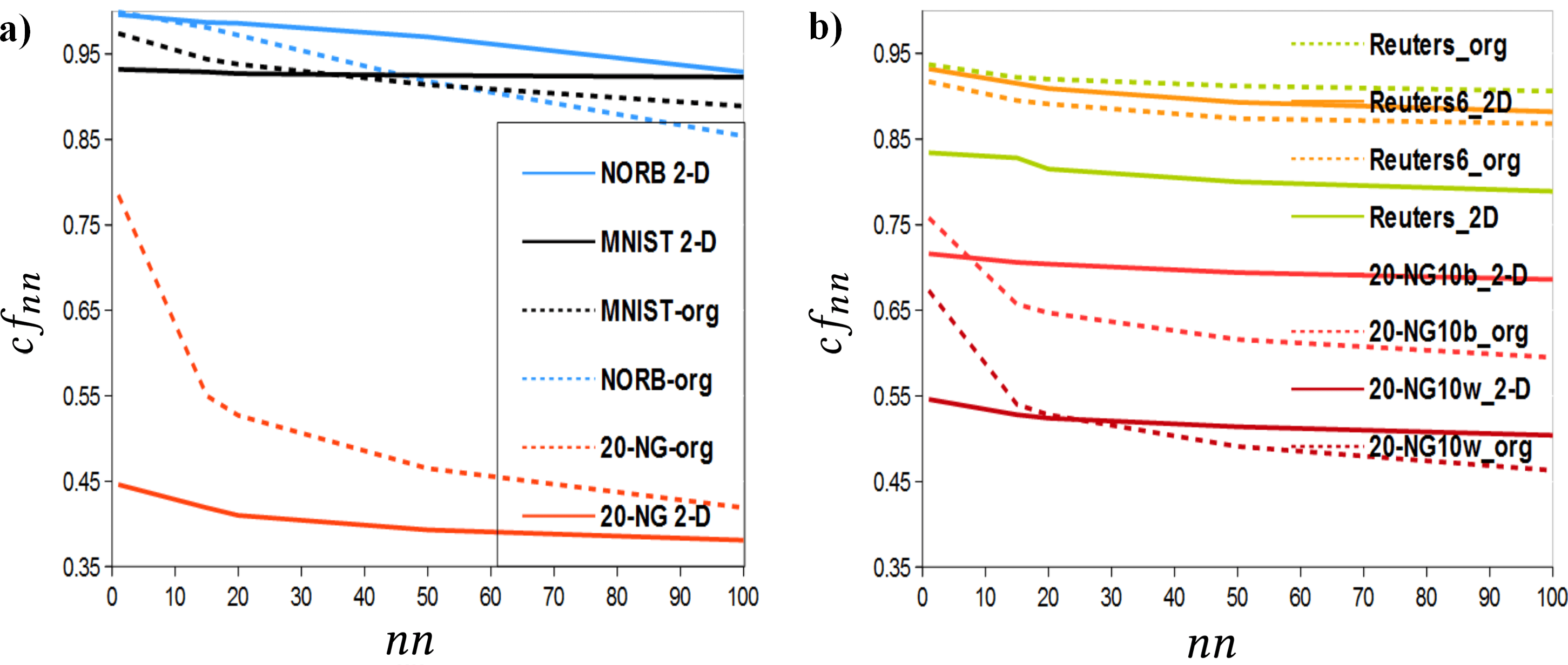}
\end{center}
\caption{The plots showing the value of $cf_{\text{\it nn}}$ with $nn$ for
various datasets for the source $\mb{Y}$ and the target $\mb{X}$ datasets.}
\label{fig:fig9}
\end{figure}

The comparison of plots from Figure~\ref{fig:fig9}, presenting the $cf_{\text{\it nn}}$ values
	with increasing $nn$ for original $N$-dimensional datasets (dashed
	lines) and their mappings (solid line), confirm high quality of
	{\bf ivhd} embeddings in terms of class separation and their purity.
The $cf_{\text{\it nn}}$ values for embeddings are more stable with $nn$ then
	original $N$-dimensional datasets. 
Only 20NG dataset, assuming all 20 classes gives rather poor embedding for BOW {\it tf-idf}
	representation (as shown in Figure~\ref{fig:fig6}, doc2vec text representation
	gives a better results). 
This is mainly due to high overlap between
	the newsgroups categories. 
Much better results are obtained
	considering 20NG embeddings for selected 10 classes (see
	Figure~\ref{fig:fig9}b).

\section{Related work}

Recently, a large body of work has been devoted to data
	embedding techniques, in the context of visual and exploratory
	analysis of high-dimensional and large datasets. 
In general, they can be roughly divided onto two groups: the classical DE
	algorithms (such as MDS, CCA, LLE, CDA, Isomap, etc.~\cite{yang2005data})
	and those based on the concept of stochastic neighbor embedding
	(SNE)~\cite{hinton2003stochastic}. 
The SNE idea can be treated as the breakthrough and
	as the milestone concept in visual analysis of large datasets.

There exists many clones of SNE in which the authors:

\begin{itemize}
\item diversify the definition of the neighborhood in the target
		and source spaces (t-SNE~\cite{maaten2008visualizing}),
\item develop more precise cost functions based on more
		complex divergence schemes (ws-SNE~\cite{yang2014optimization}),
\item combine the algorithms in hierarchical structures
		(hier\-arch\-ical-SNE~\cite{pezzotti2016hierarchical}),
\item increase their computational efficiency (bh-SNE~\cite{van2014accelerating}, qSNE~\cite{ingram2015dimensionality}, 		
		LargeVis~\cite{tang2016visualizing}, Fit-SNE~\cite{linderman2017efficient}, CE~\cite{amid2018transformation}, 
		tripletembedding~\cite{paratte2017compressive}).
\end{itemize}

All of these SNE based methods produce fantastic and very
	accurate 2-D embeddings (see, e.g., lvdmaaten.github.io/tsne/).
SNE based algorithms allow for off-line reconstruction of the
	large sets ($M\sim 10^6$) of high-dimensional ($N\sim 10^3$) data in 2-D (3-D)
	spaces with a precision not previously available. 
However, from
	the perspectives of both the big data ($M\sim 10^{7+}$) analytics and
	interactive visualization of large data ($M\sim 10^{5+}$, $N\sim 10^3$), 
	due to the high memory load and time complexity they become still too
	demanding computationally. 
As shown in~\cite{ingram2015dimensionality,pezzotti2016hierarchical,tang2016visualizing,linderman2017efficient,paratte2017compressive,dzwinel2015very}
	the efficiency of SNE algorithms can be increased, e.g., by
	employing more sophisticated data structures, and by using
	approximate nearest neighbor search methods as it is in bh-SNE
	and qSNE methods, respectively. 
However, these t-SNE clones
	suffer at least $O(M\log M)$ time complexity. The efficiency of
embedding can be further increased by easier identification of
clusters~\cite{linderman2017efficient} and by decreasing the number of calculated
distances what decreases the time complexity to $O(M)$. As shown
in~\cite{amid2018transformation}, by using a special type of data sampling, called
compressive embedding acceleration (CE), it appears that
$O(\log M)$ samples are sufficient to diffuse the information to all
$M$ data points. However, in this situation, the increase of
efficiency is at the expense of serious deterioration of embedding
quality~\cite{amid2018transformation}.

Just such the approach represents the LargeVis DE algorithm~\cite{keim2010visual}, 
which is an approximated version of t-SNE~\cite{van2014accelerating} which
exploits both DE/GV duality and data sampling. First, the
LargeVis method constructs the approximate $nn$-graph for data
prior its visualization (embedding). According to~\cite{tang2016visualizing}, this
approximated procedure should increase the efficiency of $nn$-graph construction 
assuming that its high accuracy is guaranteed.
However, as shown in our tests~\cite{klusek2017multi}, the approximated $nn$ nearest
neighbor method developed in~\cite{tang2016visualizing} is accurate only provided that
the length of $nn$ lists are at least of the same order as data
dimensionality. Otherwise, the accuracy drops rapidly (e.g. for
100-D data the accuracy drops from 95\% ($nn=200$) to 5\%
($nn=20$)). It means, that for truly high-dimensional data the $nn$
lists should be very long what considerably deteriorates the
embedding quality and decreases its computational efficiency.
This is the main reason that LargeVis results are very unstable
and need very careful parameter tuning. From the point of view
of interactive visualization of large and high-dimensional data,
also the timings reported in~\cite{tang2016visualizing,linderman2017efficient} are not encouraging.

In our position paper~\cite{dzwinel2015very}, published earlier than~\cite{tang2016visualizing}, we
reported preliminary results of the linear-time $nr$-MDS DE
method, in which we postulate using only a few the nearest $nn$
and random $rn$ neighbors $(nn + rn \sim 3\!-\!5)$. Similar to LargeVis, to
mitigate the `curse of dimensionality' effect, we assumed that all
$nn$ nearest neighbors of each data sample $\mb{y}_i$ should be kept close
to it and are equal to 0. The distances to a few $rn$ random
neighbors remain the original ones. As shown in~\cite{dzwinel2015very}, this
approximation appeared to be extremely fast giving very precise
2-D embeddings. In~\cite{dzwine2017ivhd} we demonstrated that this method is very
robust and resistant on imprecise calculation of the nearest
neighbors. Simultaneously, we developed the graph visualization
method, ivga, focused for interactive exploration of social
complex networks~\cite{dzwinel2017ivga} based on the idea from~\cite{dzwine2017ivhd}. This method
appeared to be extremely efficient and amazingly accurate~\cite{dzwinel2017ivgahist}.
It allows for reconstructing of global and local topology of big
complex networks with a few million edges on a regular laptop.
We also developed a new graph visualization technique based on,
so called B-matrices~\cite{czech2017distributed}, which merged with ivga allows for
much deeper exploration of graphs properties. In the current
paper we integrate our DE and GV visualization approaches to
develop a new, very efficient computationally, embedding
method allowing for interactive visualization of large datasets on
a laptop computer.

\section{Conclusions}

In this paper we present the method of 2-D embedding of large
	and high-dimensional data with minimal memory and
	computational time requirements. Its main concept consists in:
	increasing arbitrary the distance contrast between data vectors in
	the source dataset $\mb{Y}$ and 2-D visualization of $nn$-graph
	constructed for $\mb{Y}$. 
We have shown that substitution of the huge
	and computationally demanding distances matrix with the nearest
	neighbor $nn$-graph data structure (with small $nn$), and by
	assuming binary distance between data vectors, allow for
	decreasing radically the time and memory complexity of data
	embedding problem from $O(M\log M)$ to $O(a{\cdot} M)$ with a small value
	of $a$ coefficient. 
We also have paid attention on the theory of
	structure rigidity, which could be useful for further theoretical
	investigations, e.g., for finding minimal rigid and uniquely
	realizable graphs required for data embedding.

We have demonstrated that {\bf ivhd} outperforms, in terms of
	computational time, the state-of-the-art DE algorithms more than one
	order of magnitude on the standard DE benchmark data. 
Data embeddings obtained by {\bf ivhd} are also very effective in reconstructing
	data separation in high-dimensional and large datasets. 
This is the principal requirement for knowledge mining, because just the
	multi-scale clusters of data represent basic `granules of
	knowledge'. 
Due to simplicity of {\bf ivhd} algorithm -- it is in fact a
	clone of the classical MDS algorithm -- its efficiency can be
	further increased by implementing its parallel versions in GPU
	and MIC environment, what we did already with classical MDS
	method (see~\cite{pawliczek2013interactive,pawliczek2014visual,pawliczek2015visual}).

It is worth to mention, that {\bf ivhd} yields less impressive results
	than recently published SNE clones in reconstructing the local
	neighborhood. 
The `crowding problem', similar to that in SNE,
	causes that the most of datapoints in 2-D embeddings gather in the
	center of clusters. 
Nevertheless, the accurate
	reconstruction of the nearest neighbors is rather the secondary
	requirement in interactive visualization of big data. 
The exact $nn$ lists (or their approximations) are well known prior to data
	embedding. 
They can be presented anytime visually, e.g., as the datapoints' spines in $\mb{X}$.
Moreover, the exact lists of the nearest neighbors are not reliable
	from the context of both measurement errors and the `curse of
	dimensionality' principle. 
Besides, if more accurate
	reconstruction of data locality would be required, the {\bf ivhd} can be
	used for generation of initial configuration for much slower SNE
	based DE algorithms. 
Unlike SNE competitors, {\bf ivhd}
	has only three free parameters which should be fit to data: $nn$, $rn$ and
	$c$. 
Because the method is fast, they can be easily matched
	interactively, although `universal' set, i.e, $nn=3$, $rn=1$ and $c=0.1$
	(or $c=0.01$), worked very well for most of data (and networks)
	visualized by the Authors of this paper. 
Summing up, we highly
	recommend {\bf ivhd} as a good candidate for efficient data
	embedding engine in interactive visualization of truly big data.
The recent Win32 version of our software is available here: https://goo.gl/6HzEd3.



\bibliographystyle{acm}
\bibliography{DzwinelWcisloMatwin2dEmbedding}

\end{document}